\documentclass[floatfix,rmp,twocolumn,twoside]{revtex4}
\usepackage{graphicx}

\bibpunct{[}{]}{,}{n}{}{,}

\usepackage[english]{babel}
\usepackage{verbatim}
\usepackage{varwidth}
\usepackage{graphicx}
\usepackage{alltt}

\makeatletter
\g@addto@macro\@verbatim\small
\makeatother

\begin{document}

\title{A Constraint-Satisfaction Parser for Context-Free Grammars}
\author{Luis~Quesada, Fernando~Berzal, and Francisco J.~Cortijo\\
  Department of Computer Science and Artificial Intelligence, CITIC, University of Granada, \\
  Granada 18071, Spain \\
  \textit{lquesada@decsai.ugr.es, fberzal@decsai.ugr.es, cb@decsai.ugr.es}
  }

\begin{abstract}
Traditional language processing tools constrain language designers to specific kinds of grammars.
In contrast, model-based language specification decouples language design from language processing. As a consequence, model-based language specification tools need general parsers able to parse unrestricted context-free grammars.
As languages specified following this approach may be ambiguous, parsers must deal with ambiguities.
Model-based language specification also allows the definition of associativity, precedence, and custom constraints. Therefore parsers generated by model-driven language specification tools need to enforce constraints.
In this paper, we propose Fence, an efficient bottom-up chart parser with lexical and syntactic ambiguity support that allows the specification of constraints and, therefore, enables the use of model-based language specification in practice.
\end{abstract}

\maketitle

\section{Introduction}
Traditional language specification techniques \cite{Aho1972} require the developer to provide a textual specification of the language grammar.

In contrast, model-based language specification techniques \cite{Quesada2011b} allow the specification of languages by means of data models annotated with constraints. 

Model-based language specification has direct applications in the following fields: programming tools \cite{Aho2006}, domain-specific languages \cite{Fowler2010,Hudak1996,Mernik2005}, model-driven software development \cite{Schmidt2006}, data integration \cite{Tan2006}, text mining \cite{Turmo2006,Crescenzi2004}, natural language processing \cite{Jurafsky2009}, and the corpus-based induction of models \cite{Klein2004}.

Due to the nature of the aforementioned application fields, the specification of separate language elements may cause lexical and syntactic ambiguities.
Lexical ambiguities occur when an input string simultaneously corresponds to several token sequences \cite{Nawrocki1991}, which may also overlap.
Syntactic ambiguities occur when a token sequence can be parsed in several ways.

The formal grammars of languages specified using model-based techniques may contain epsilon productions (such as $E := \epsilon$), infinitely recursive production sets (such as $A := c$, $A := B$, and $B := A$), and associativity, precedence, and custom constraints.
Therefore, a parser that supports such specification is needed.

Our proposed algorithm, Fence, is a bottom-up chart parser that accepts a lexical analysis graph as input, performs an efficient syntactic analysis taking constraints into account, and produces a parse graph that represents all the possible parse trees.
The parsing process discards any sequence of tokens that does not provide a valid syntactic sentence conforming to the language specification, which consists of a production set and a set of constraints.
Fence implicitly performs a context-sensitive lexical analysis, as the parsing process determines which token sequences end up in the parse graph.
Fence supports every possible construction in a context-free language with constraints, including epsilon productions and infinitely recursive production sets.

The combined use of the Lamb lexical analyzer \cite{Quesada2011a} and Fence allows the generation of processors for languages with ambiguities and constraints, and it renders model-based language specification techniques feasible. Indeed, ModelCC \cite{Quesada2011b} is a model-based language specification tool that relies on Lamb and Fence to generate language processors.

\section{Background} \label{sec:sec2}

Language processing tools traditionally divide the analysis into two separate phases; namely, scanning (or lexical analysis), which is performed by lexers, and parsing (or syntax analysis), which is performed by parsers.
However, language processing tools based on scannerless parsers also exist.

\subsection{Lexical Analysis Algorithms with Ambiguity Support}

Given a language specification describing the tokens listed in Figure \ref{fig:tokens}, the string ``\&5.2\& /25.20/'' can correspond to the four different lexical analysis alternatives shown in Figure \ref{fig:analysis}, depending on whether the sequences of digits separated by points are considered real numbers or integer numbers separated by points.

\begin{figure}[b]
\centering
\begin{varwidth}{\linewidth}
\begin{verbatim}
(-|\+)?[0-9]+           Integer
(-|\+)?[0-9]+\.[0-9]+   Real
\.                      Point
\/                      Slash
\&                      Ampersand
\end{verbatim}
\end{varwidth}
\caption{Specification of token types as regular expressions for a lexically-ambiguous language.}
\label{fig:tokens}
\end{figure}

\begin{figure}[t]
\begin{itemize}
\item \texttt{\fontsize{2.95mm}{2.95mm}\selectfont Ampersand Integer Point Integer Ampersand Slash Integer Point Integer Slash}
\item \texttt{\fontsize{2.95mm}{2.95mm}\selectfont Ampersand Integer Point Integer Ampersand Slash Real Slash}
\item \texttt{\fontsize{2.95mm}{2.95mm}\selectfont Ampersand Real Ampersand Slash Integer Point Integer Slash}
\item \texttt{\fontsize{2.95mm}{2.95mm}\selectfont Ampersand Real Ampersand Slash Real Slash}
\end{itemize}
\caption{Different possible token sequences in the input string ``\&5.2\& /25.20/'' due to the lexically-ambiguous language specification shown in Figure \ref{fig:tokens}.}
\label{fig:analysis}
\end{figure}

\begin{figure}[htb!]
\centering
\begin{varwidth}{\linewidth}
\begin{verbatim}
E ::= A B
A ::= Ampersand Real Ampersand
B ::= Slash Integer Point Integer Slash
\end{verbatim}
\end{varwidth}
\caption{Context-sensitive productions that resolve the ambiguities in Figure \ref{fig:analysis}.}
\label{fig:srules1}
\end{figure}

\begin{figure*}[p]
\centering
\includegraphics[scale=0.385]{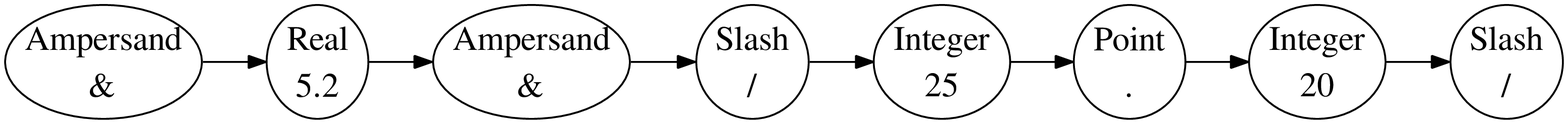}
\caption{Desired lexical analysis of the lexically ambiguous ``\&5.2\& /25.20/'' input string.}
\label{fig:e4}
\end{figure*}

The productions shown in Figure \ref{fig:srules1} illustrate a scenario of lexical ambiguity sensitivity.
Sequences of digits separated by points should be considered either \emph{Real} tokens or \emph{Integer Point Integer} token sequences depending on the surrounding tokens, which may be either \emph{Ampersand} tokens or \emph{Slash} tokens.
The desired result of analyzing the input string ``\&5.2\& /25.20/'' is shown in Figure \ref{fig:e4}.

The further application of a parser supporting lexical ambiguities would produce the only possible valid sentence, which, in turn, would be based on the only valid lexical analysis for our example.
The intended results are shown in Figure \ref{fig:e5}.

\begin{figure*}[p]
\centering
\noindent \includegraphics[scale=0.385]{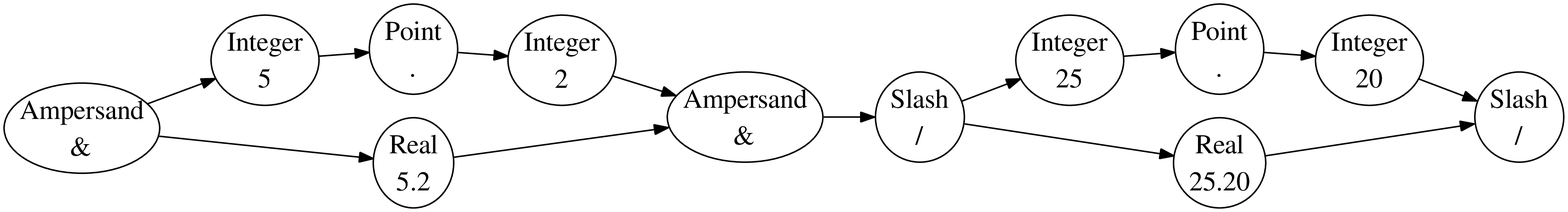}
\caption{Lexical analysis graph, as produced by the Lamb lexer.}
\label{fig:e1}
\end{figure*}

\begin{figure*}[p]
\centering
\includegraphics[scale=0.385]{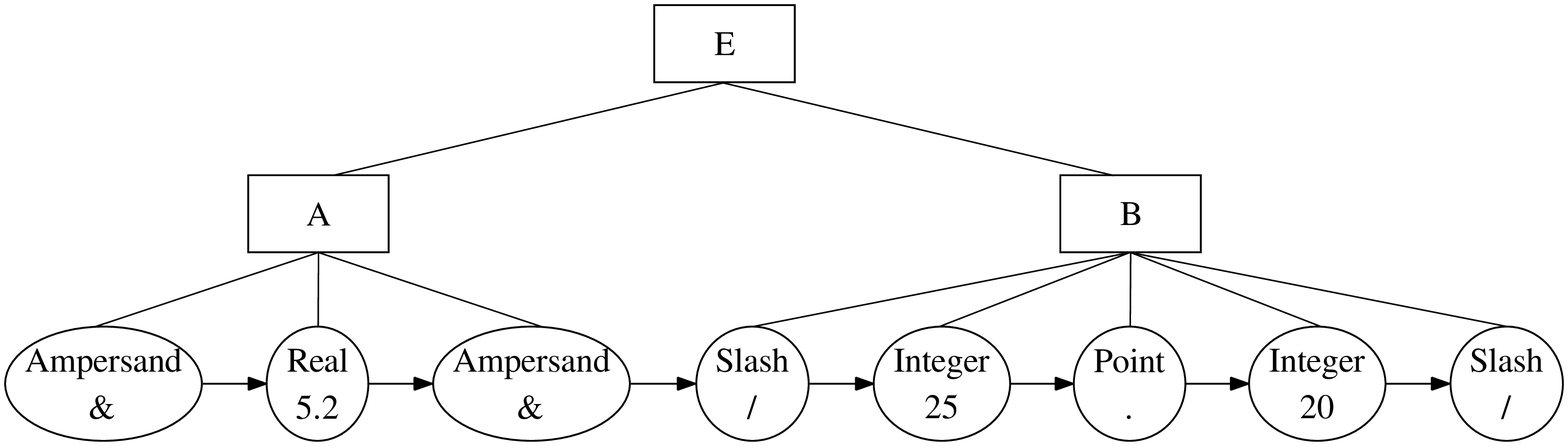}
\caption{Syntactic analysis graph, as produced by applying a parser that supports lexical ambiguities to the lexical analysis graph shown in Figure \ref{fig:e1}.
Squares represent nonterminal symbols found during the parsing process.}
\label{fig:e5}
\end{figure*}

\begin{figure*}[p]
\centering
\includegraphics[scale=0.385]{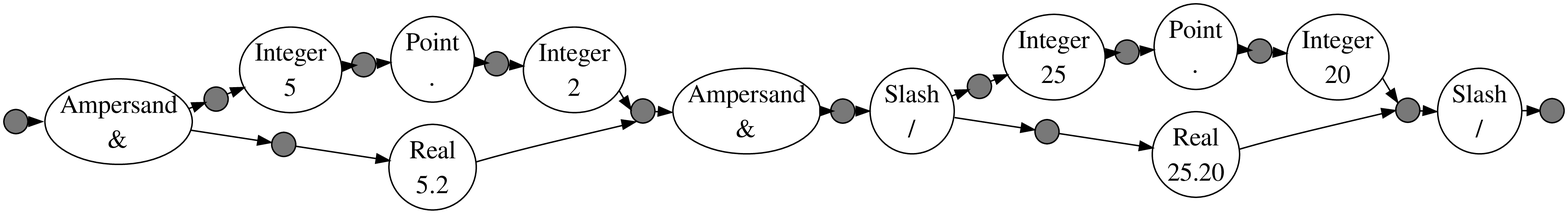}
\caption{Extended lexical analysis graph corresponding to the lexical analysis graph shown in Figure \ref{fig:e1}. Gray nodes represent cores.}
\label{fig:e6}
\end{figure*}

\begin{figure*}[p]
\centering
\includegraphics[scale=0.385]{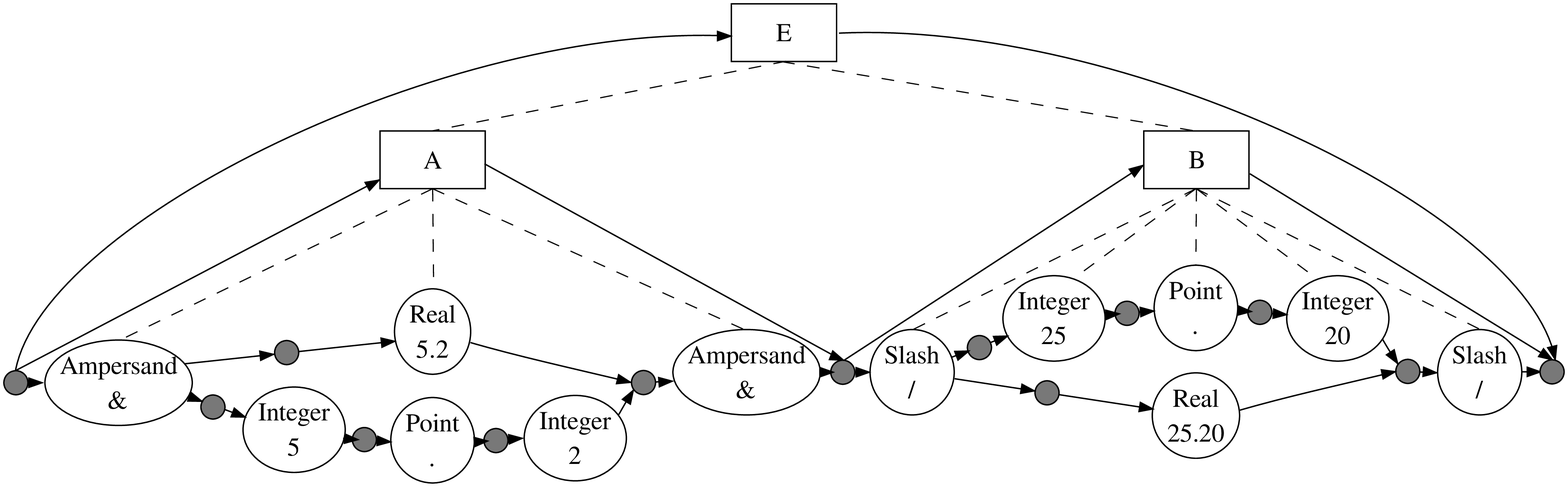}
\caption{Parse graph corresponding to the extended lexical analysis graph shown in Figure \ref{fig:e6}. 
Squares represent nonterminal symbols found during the parsing process. Dotted lines represent the explicit parse graph node.}
\label{fig:e7}
\end{figure*}

The Lamb lexical analyzer \cite{Quesada2011a} captures all possible sequences of tokens within a given input string and it generates a lexical analysis graph that describes them all, as shown in Figure \ref{fig:e1}.
In these graphs, each token is linked to its preceding and following tokens. There may also be several starting tokens.
Each path in these graphs describes a possible sequence of tokens that can be found within the input string.

To the best of our knowledge, the only way to process lexical analysis graphs consists of extracting the different paths from the graph and parse each of them.
This process is inefficient, as partial parsing trees that are shared among different token sequences have to be created several times.

\subsection{Syntactic Analysis Algorithms}

Traditional efficient parsers for restricted context-free grammars, such as the LL \cite{Oettinger1961}, LR \cite{Knuth1965}, LALR \cite{DeRemer1969,DeRemer1982}, and SLR \cite{DeRemer1971} parsers, do not consider ambiguities in syntactic analysis, so they cannot be used to parse ambiguous languages. The efficiency of these parsers is $O(n)$, being $n$ the token sequence length.

Generalized LR (GLR) parsers \cite{Lang1974} parse in linear to cubic time, depending on how closely the grammar conforms to the underlying LR strategy. The time required to run the algorithm is proportional to the degree of nondeterminism in the grammar. The Universal parser \cite{Tomita1987} is a GLR parser used for natural language processing. However, it fails for grammars with epsilon productions and infinitely recursive production sets.

Existing chart parsers for unrestricted context-free grammar parsing, as the CYK parser \cite{Younger1967,Kasami1969} and the Earley parser \cite{Earley1983}, can consider syntactic ambiguities but not lexical ambiguities. The efficiency of these general context-free grammar parsers is $O(n^3)$, being $n$ the token sequence length.

\section{Fence}

In this paper, we introduce Fence, an efficient bottom-up chart parser that produces a parse graph that contains as many root nodes as different parse trees exist for a given ambiguous input string.

In contrast to the parsing techniques mentioned in the previous section, Fence is able to process lexical analysis graphs and, therefore, it efficiently considers lexical ambiguities.

Fence also considers syntactic ambiguities, allows the specification of constraints, and supports every possible context-free language construction, particularly epsilon productions and infinitely recursive production sets.

The Fence parsing algorithm consists of three consecutive phases: the extended lexical analysis graph construction phase, the chart parsing phase, and the constraint enforcement phase.

\subsection{Terminology}

A context-free grammar is formally defined \cite{Chomsky1956} as the tuple $(N,\Sigma,P,S)$, where:
\begin{itemize}
\item $N$ is the finite set of nonterminal symbols of the language, sometimes called syntactic variables, none of which appear in the language strings.
\item $\Sigma$ is the finite set of terminal symbols of the language, also called tokens, which constitute the language alphabet
(i.e. they appear in the language strings). Therefore, $\Sigma$ is disjoint from $N$.
\item
$P$ is a finite set of productions, each one of the form $N \rightarrow (\Sigma \cup N)^{*}$, where $*$ is the Kleene star operator, $\cup$ denotes set union, the part before the arrow is called the left-hand side (LHS) of the production, and the part after the arrow is called the right-hand side (RHS) of the production.
after the arrow is called the right-hand side of the production.
\item $S$ is a distinguished nonterminal symbol, $S \in N$: the grammar start symbol.
\end{itemize}

A dotted production is of the form $N \rightarrow (\Sigma \cup N)^{*}.(\Sigma \cup N)^{*}$, where the dot indicates that the RHS symbols before the dot have already been matched with a substring of the input string.

A handle is a tuple $(dotted production,[start,end])$, where $start$ and $end$ identify the substring of the input string that matched the dotted production RHS symbols before the dot.
Each handle can be used during the parsing process to match a rule RHS symbol with a node representing either a token or a nonterminal symbol (namely, SHIFT actions in LR-like parsers) or perform a reduction (namely, REDUCE actions in LR-like parsers).

A core is a set of handles.

\subsection{Extended Lexical Analysis Graph Construction Phase} \label{sec:extended}

In order to efficiently perform the parsing process, Fence converts the input lexical analysis graph (LA graph) into an extended lexical analysis graph (ELA graph) that stores information about partially applied productions (namely, handles) in data structures (namely, cores).

In an ELA graph, tokens are not linked to their preceding and following tokens, but to their preceding and following cores.
Cores are, in turn, linked to their preceding and following token sets.
For example, the ELA graph corresponding to the LA graph in Figure \ref{fig:e1} is shown in Figure \ref{fig:e6}.

The conversion from the LA graph to the ELA graph is performed by completing the LA graph with cores.
A \emph{starting} core is linked to the tokens with an empty preceding token set.
A \emph{last} core is linked from the tokens with an empty following token set.
Finally, for each one of the other tokens in the LA graph, a preceding core is linked to it.
Links between tokens in the LA graph are converted into links from tokens to the cores preceding each token of their following token set in the ELA graph.

\subsection{Chart Parsing Phase} \label{sec:parsing}

The Fence chart parsing phase processes the ELA graph and generates an implicit parse graph (I-graph).
Nodes in the I-graph are described as $(start,end,symbol)$ tuples, where $start$ and $end$ identify the substring of the input string, and $symbol$ identifies the production LHS.
It should be noted that ambiguities, both lexical and syntactic, are implicit in the I-graph nodes, as they contain no information about their contents.
The I-graph contains a set of starting nodes, each of which may represent several parse tree roots.
The parsing itself is performed by progressively applying productions and storing handles in cores.

The grammar productions with an empty RHS (i.e. epsilon productions) are removed from the grammar and their LHS symbol is stored in the \emph{epsilonSymbols} set.
This set allows these parse symbols being skipped when found in a production, as if a reduction using the epsilon production were applied.

The agenda is a stack of $(handle,node)$ in which the node can match the symbol after the dot in the dotted rule of the handle. It is initially empty.

The \emph{alreadyGenerated} handle set contains all the agenda entries ever generated and inhibits the generation of duplicate entries.

The parser is initialized by generating a handle for each production and adding them to every core, as shown in Figure \ref{fig:codeinit}. 

The \emph{addHandle} procedure in Figure \ref{fig:codeaddrule} is responsible for adding a handle to a core. It also adds the corresponding agenda entries for that handle with the nodes that follow the core and match the symbol after the dot in the dotted production of the handle.
It should be noted that the \emph{addHandle} procedure considers epsilon productions: if a production RHS symbol is in the \emph{epsilonSymbols} set, both the possibilities of it being reduced or not by that production are considered; that is, a new handle that skips that element is added to the same core. 
It should also be noted that element are skipped iteratively, as many consecutive RHS symbols of a production could be in the \emph{epsilonSymbols} set.

\begin{figure}[htb]
\begin{verbatim} 
procedure addHandle(Production p, int matched,
              ImplicitNode first,ImplicitNode n,
              Stack<[Handle,ImplicitNode]> agenda):
  offset = 0
  do:
    next = matched+offset
    nextSymbol = p.right[next].symbol
    h = new Handle(p,next,first,first.startIndex)
    if !n.core.contains(h):
      n.core.add(h)
    if n.symbol == nextSymbol:
      if !alreadyGenerated.contains([h,n]):
        agenda.push([h,n])
        alreadyGenerated.add(]h,n])
    offset++
  while epsilonSymbols.contains(nextSymbol) &&
        next<r.right.size
\end{verbatim}
\caption{Pseudocode of the ancillary \emph{addHandle} procedure.}
\label{fig:codeaddrule}
\end{figure}

\begin{figure}[htb]
\begin{verbatim}
agenda = {}
for each Production p in productionSet:
  for each ImplicitNode n in nodeSet:
    addHandle(p,0,n,n,agenda)
\end{verbatim}
\caption{Pseudocode of the chart parser initialization.}
\label{fig:codeinit}
\end{figure}

\begin{figure}[htb]
\begin{verbatim}
  while !agenda.empty:
    [h,n] = agenda.pop()
    if h.dotposition == h.production.right.size-1:
      // Production matched all its elements.
      // i.e. Reduction
      nn = new ImplicitNode(h.startIndex,
                            n.endIndex,
                            p.left.symbol)
      h.first.core.following.add(nn)
      nn.preceding.add(h.first.core)
      for each Core c in n.following:
        c.preceding.add(nn)
        nn.following.add(c)
      for each Handle hn in
                h.first.core.waitingFor(nn.symbol):
        hadd = new Handle(hn.production,hn.next,
                          hn.first,hn.startIndex)
        agenda.push([hadd,n])
    else:
      // i.e. Shift
      for each Core c in n.following:
        for each ImplicitNode nnext in c.following:
          addHandle(h.production,h.next+1,h.first,
                    h.startIndex,agenda)
\end{verbatim}
\caption{Pseudocode of the Fence parsing phase.}
\label{fig:codeprocess}
\end{figure}

The parsing process consists in iteratively extracting entries consisting of handles and nodes from the agenda and matching the next symbol of the RHS of the handle production with the node.
The handles whose productions are successfully matched are added to the cores following the node and the agenda is updated with the entries that contain any of the newly generated handles.
In case all the symbols of a production RHS match a sequence of nodes, a new node is generated by reducing them.
The new $node$ start index is obtained from the handle, its $end$ position is obtained from the last node matched, and its $symbol$ is the LHS symbol of the production.
When a newly generated node only has the \emph{starting} core in its preceding core set and the \emph{final} core in its following core set, and its $symbol$ corresponds to the initial symbol of the grammar, it is added to the parse graph starting node set, which means that that node represents a valid parse.
The pseudocode for this process is shown in Figure \ref{fig:codeprocess}.

The result of the chart parsing phase is an I-graph, which the constraint enforcement phase accepts as input.

The Fence chart parsing phase order of efficiency is theoretically equivalent to existing Earley chart parsers. That is, $O(n^3)$ in the general case, $O(n^2)$ for unambiguous grammars, and $O(n)$ for almost all LR(k) grammars, being $n$ the length of the input string.

\subsection{Constraint Enforcement Phase} \label{sec:constraint}

The Fence constraint enforcement phase processes the I-graph and generates an explicit parse graph (E-graph, or just parse graph) by enforcing the constraints defined for the language.
Nodes in the E-graph that represent tokens are still defined as $(start,end,symbol)$ tuples.
Nodes in the E-graph that represent nonterminal symbols reference the list of nodes that matched the production used to generate those nodes.
It should be noted that ambiguities, both lexical and syntactic, are explicit in the E-graph, as it represents several parse trees corresponding to all the possible interpretations of the input string.
The E-graph contains a set of starting nodes, each of which represents a parse tree root.
Constraint enforcement is performed by converting each implicit node into every possible explicit node sequence that can be derived from the implicit node and satisfies the specified constraints; that is, by expanding the each implicit node.

Only the nodes that conform valid parse trees are needed in the parse graph. In order to generate only these nodes, each one of the implicit nodes in the starting node set of the I-graph is recursively expanded using memoization. 
Each possible resulting explicit node is the root of a parse tree in the E-graph.

\subsubsection{Algorithm Description}

The expansion of an implicit node is performed by finding every possible reduction of a sequence of explicit nodes that generates that node.
Each one of these reductions produces an explicit node.
Whenever an implicit node is found and needed in order to make the reductions progress, it is expanded recursively.
It should be noted that this procedure is different from parsing itself in that the actual bounds of the reductions for every node are known.

The \emph{expand} procedure in Figure \ref{fig:codeexpand} expands an implicit node by applying every possible production that could generate it and produces a set of explicit nodes.
The use of the \emph{history} set inhibits entering an infinite loop when processing infinitely recursive production sets, as it avoids the expansion of a node as an indirect requirement of expanding the same node.

\begin{figure}[htb]
\begin{verbatim}
procedure expand(ImplicitNode n,
                 Set<ImplicitNode> history,
                 Map<ImplicitNode,
                 Set<Node>> alreadyExpanded)
                 returns Set<Symbol>:
  if alreadyExpanded.contains(n): // memoization
    return alreadyExpanded.get(n)
  else:
    // the history set avoids infinite loop in
    // recursive production sets
    if !history.contains(n):
      history.add(n);
      // try to apply every production
      for each Production p with
                          LHS symbol == n.symbol:
        for every ImplicitNode pn with
                        startIndex == n.startIndex:
          if pn != n && pn.endIndex<=n.endIndex:
            if p.mayMatch(pn.symbol):
              // apply production p to each
              // expanded symbol of pn
              pn.expandeds = expand(pn,history,
                                   alreadyExpanded)
              for each Node nn in pn.expandeds:
                out += apply(p,nn,0,{},
                           alreadyExpanded,history)
    alreadyExpanded.put(n,out)
    return out
\end{verbatim}
\caption{Pseudocode of the \emph{expand} procedure that obtains every possible derivation of a given node in the parse graph.}
\label{fig:codeexpand}
\end{figure}

\begin{figure}[htb]
\begin{verbatim}
procedure apply(Production p, Node n, int matched,
       List<Node> content,
       Map<ImplicitNode,Set<Node>> alreadyExpanded,
       Set<ImplicitNode> history)
       returns Set<Node>:
  if matched == p.right.size:
    n = new Node(p.symbol,p,content)
    if checkConstraints(n):
      return {n}
  else:
    offset = 0
    next = matched+offset
    do:
      if p.right[next].symbol == n.symbol:
        for each ImplicitNode pn in
                                n.followingNodes():
          if pn is the next symbol to match
                                 in the production:
            // keep applying production to each
            // expanded symbol of pn
            expandeds = expand(pn,history,
                                   alreadyExpanded)
            for each Node nn in expandeds:
              out += apply(p,nn,next+1,content+n,
                           alreadyExpanded,history)
      offset++
      next = matched+offset
    while epsilonSymbols.contains(nextSymbol) && 
                 next<r.right.size &&
                 p.right[next].symbol == n.symbol
    return out
\end{verbatim}
\caption{Pseudocode of the ancillary \emph{apply} procedure that applies a production.}
\label{fig:codeapply}
\end{figure}

The \emph{apply} procedure in Figure \ref{fig:codeapply} applies a production by matching the RHS symbol given by the $matched+1$ index of it with the $n$ node, expanding the nodes that follows it, and recursively applying the next RHS symbols of the production.

The \emph{checkConstraints} procedure is the responsible for the enforcement of the constraints specified by the developer.

\subsubsection{Supported Constraints}

Fence supports associativity constraints, selection precedence constraints, composition precedence constraints, and custom-designed constraints.

The fact that the constraint check is performed during the graph expansion improves the parser performance, as the sooner constraints are applied, the more interpretations are discarded.
For example, in the case of a binary expression with left-to-right associative operators, the string ``2+5+3+5+6+2+1+5+6+3'' can be expanded in $10!$ possible ways without considering the associativity constraint, and in just $1$ possible way when considering it.

\begin{itemize}
\item {\bf Associativity constraints} allow the specification of the associative property for binary operators.
The application of a production is inhibited when one of the nodes that matches its RHS symbols has an associativity constraint and is followed (for left-to-right associativity constraints), preceded (for right-to-left associativity constraints), or either followed or preceded (for non-associative associativity constraints) by a node that was derived using the same production.

\item {\bf Selection precedence constraints} allow the resolution of syntactic ambiguities caused by different explicit nodes (i.e. interpretations) resulting from a single implicit node.
For example, a \emph{Statement} can be either an \emph{OutputStatement} or a \emph{FunctionCall}. Both \emph{OutputStatement} and \emph{FunctionCall} can match the input string ``output(var);'', therefore \emph{OutputStatement} can be set to precede \emph{FunctionCall}, which will inhibit that string from being considered a function call.
The application of a production is inhibited when it is preceded by a different production and both of them match the same node sequence.

\item {\bf Composition precedence constraints} allow the resolution of syntactic ambiguities when a node derived using a production cannot be derived using another production.
For example, one of the productions \emph{ConditionalStatement ::= ``if'' Expression Sentence} and \emph{ConditionalStatement ::= ``if'' Expression Sentence ``else'' Sentence} can be set to precede the other one in order to resolve the ambiguity in ``if expr1 if expr2 sent1 else sent2'', in which ``else sent2'' could be assigned to either the inside or outside conditional sentence.
The application of a production is inhibited when it precedes any of the productions used to derive the nodes that matched its RHS symbols. 

\item {\bf Custom-designed constraints} allow the specification of any other constraints (e.g. semantic constraints).
In order to enforce custom-designed constraints, an evaluator can be assigned to any production. Whenever a node is generated, the evaluator of the production used to derive it gets executed and determines whether the node satisfies the constraint or not. In the later case, its generation is inhibited.
Custom-designed constraints provide a very extensible framework which allows developers to design complex syntactic or semantic constraints (e.g. probabilistic constraints, corpus-based constraints) that effectively limit the possible interpretations of an input string and, as a side effect, improve the performance of the parser, as pruned partial interpretations are discarded as soon as they do not fulfill the constraints.
\end{itemize}

The result of the constraint enforcement step is an E-graph or parse graph, such as the one shown in Figure \ref{fig:e7}.

The Fence constraint enforcement phase improves the performance of traditional techniques phases in practice, as all constraints are applied at the earliest possible time, thus discarding possibilities that would otherwise be processed later.

\section{Conclusions and Future Work}

We have presented Fence, an efficient bottom-up chart parsing algorithm with lexical and syntactic ambiguity support. Its constraint-based ambiguity resolution mechanism enables the use of model-based language specification in practice. In fact, the ModelCC model-based language specification tool \cite{Quesada2011b} generates Fence parsers.

Fence accepts a lexical analysis graph as input, performs syntactic analysis conforming to a formal context-free grammar specification and a set of constraints, and produces as output a compact representation of the set of parse trees accepted by the language.

Fence applies constraints while expanding the parse graph. Thus, it improves the performance of traditional techniques in practice, as the sooner constraints are applied, the less processing time and memory the parser will require.

In the future, we plan to apply the ModelCC model-based language specification tool, which relies on Fence, to the implementation of programming tools, model-driven software development, data integration, corpus-based induction of models, text mining, and natural language processing.



\bibliographystyle{plain}
\bibliography{doc}

\end{document}